\documentclass{article}
\usepackage{spconf,amsmath,graphicx,hyperref,xcolor,multirow}
\usepackage{enumitem}
\usepackage[table,xcdraw]{xcolor}
\usepackage[ruled,linesnumbered]{algorithm2e}
\setlist[itemize]{noitemsep,
                  topsep=0pt,
                  parsep=0pt,
                  itemsep=0pt}

\title{RoiMAM: Region-of-Interest Medical Attention Model for Efficient Vision-Language Understanding}
\name{Jiayan Yang$^{1*}$\thanks{*Work done during the author's internship at the Shenzhen Institutes of Adavance Techonology. \dag Correspondence: wq.fang@siat.ac.cn}, Zhuoyu Wu$^{1,2}$, Wenqi Fang$^{1\dag}$\thanks{W.F. was supported by National Natural Science Foundation of China (NSFC) under Grant No.12401676.}}
\address{$^1$Shenzhen Institutes of Advanced Technology, Chinese Academy of Science  \\
$^2$CyPhi($\Psi\Phi$) AI Research Lab, School of IT, Monash University, Malaysia Campus}

\begin{document}
%
\maketitle
\begin{abstract}
Vision–Language Models (VLMs) facilitate medical visual question answering (MedVQA) by jointly interpreting images and text. However, existing models typically depend on large architectures and closed-set answers, which limits their efficiency and potential clinical applicability. To overcome these shortcomings, we introduce RoiMAM, an efficient VLM. It integrates a training-free ROI Generation Module with Semantic Selective Suppression to focus on lesion-relevant regions, alongside a Text Prompt Enhancer module that provides modality-specific context without introducing training parameters. Compared to the widely used MedVInT-TD model, our design achieves efficient and accurate diagnosis at less than 20\% of the model size, while improving accuracy by approximately 2\% on SLAKE and 4.6\% on PMC-VQA.
\end{abstract}
\section{INTRODUCTION}
\label{sec:intro}


Computer-aided medical diagnosis has progressed rapidly with the advent of deep-learning technologies. Recent Vision–Language Models (VLMs), usually built on Large Language Models (LLMs), have advanced medical visual question answering (MedVQA) systems by jointly interpreting clinical images and textual queries. These systems show promise for addressing healthcare challenges such as resource shortages, physician workload, and diagnostic errors.

Despite notable progress, MedVQA systems still face significant barriers to practical adoption. Most existing VLMs remain confined to closed-set answering—selecting from predefined options rather than generating free-form responses—creating a mismatch with the open-ended nature of clinical consultations \cite{llavamed}. Although LLMs exhibit strong conversational ability, their heavy computational overhead and large parameter counts make training and inference inefficient, limiting scalability and delaying real-world implementation \cite{efficient}.

These scalability concerns have driven research into techniques that enhance model capability without incurring heavy computational cost. A promising avenue is the use of visual prompts to guide networks toward lesion-relevant regions in medical images. Yet, these region-of-interest (ROI)–based prompting approaches still rely on supervised detection or segmentation models, which demand expensive, large annotated datasets \cite{FAVP}\cite{medvp}. Consequently, their pipelines remain resource-intensive, and hallucinations in lesion localization further erode the reliability required for clinical deployment.

To overcome these limitations, we present \textbf{RoiMAM}, an ultra-lightweight and efficient VLM with only 1.7 billion (B) parameters that delivers state-of-the-art results on multiple MedVQA benchmarks while maintaining both accuracy and computational efficiency. RoiMAM combines a training-free ROI Generation Module (RGMo), which uses the Semantic Selective Suppression ($S^3$) mechanism to emphasize text-relevant image regions, with a Text Prompt Enhancer (TPE) module that adds modality-specific context without extra parameters. 

The main contributions of this study are outlined below:
\begin{itemize}
    \item To alleviate the computational burden of traditional VLMs in the MedVQA tasks, we propose an efficient framework called RoiMAM, which integrates RGMo and TPE modules to enhance performance while reducing parameters. 
    \item 
     A training-free RGMo module is proposed, employing the $S^3$ mechanism to minimize irrelevant visual attention and enable accurate lesion localization with minimal computation and no pre-training.
    
    \item We introduce the TPE module, leveraging existing vision encoders to enrich text context and improve cross-modal robustness without adding parameters.
    
    \item 
    Our model outperforms large models such as MedVInT-TD on three representative open-source datasets, obtaining higher accuracy while reducing the number of parameters by roughly 80\%, thus mitigating data privacy and computational constraints\cite{pmcvqa}.
    
\end{itemize}

\section{RELATED WORKS}
MedVQA systems aim to answer questions from medical images, mirroring real-world diagnostic reasoning. Early works, mainly based on the CLIP-based model, framed MedVQA as a classification task with predefined answer sets. For instance, PubMedCLIP demonstrated CLIP’s applicability to the medical domain, and later datasets like PMC-VQA and PMC-OA facilitated larger-scale pretraining of VLMs \cite{pmcvqa}\cite{pubmedclip}\cite{pmc-clip}. While effective for closed-set tasks, these approaches do not fully reflect clinical practice, where diagnoses are open-ended and answers are not limited to fixed options.

With the rise of generative LLMs such as
LLaMA and their visual extensions, VLMs
are being increasingly adapted for
medical applications. HuatuoGPU-Vision fine-tuned VLMs on task-specific datasets, while LLaVA-Med introduced a dedicated medical dataset and a biomedical multimodal instruction-following pipeline  \cite{llavamed}\cite{huatuo}. In addition, visual prompts have also been explored: FAVP proposed an Adaptive Visual Prompt Creator (AVPC) \cite{FAVP}\cite{sammed2d}, and MedVP utilized a fine-tuned Grounding DINO to locate ROIs from question keywords \cite{medvp}. However, these methods remain unsuitable for on-device deployment because of their large model sizes, and pretraining visual prompt generators is still computationally costly due to the need for annotated images. In contrast, our RoiMAM model provides a training-free visual prompt generation method to pinpoint relevant regions. This keeps the parameter size of RoiMAM below two billion, making it feasible for future edge deployment.

\begin{figure*}
    \centering
    \centerline{\includegraphics[width=18cm]{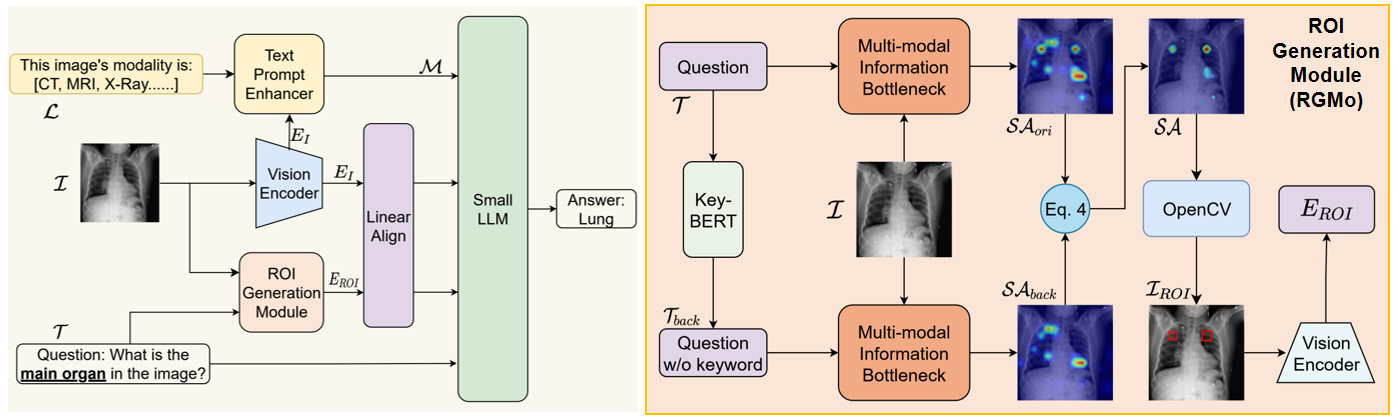}}
    \caption{\textbf{Left:} Overview of the RoiMAM framework. 
    \textbf{Right:} The workflow of the RGMo module. 
    }
    
    \label{fig:procedure}
\end{figure*}

\section{PROPOSED METHOD}
\label{sec:method}


In this section, we detail our RoiMAM framework, designed to process and synthesize information from three distinct streams: an input image ($\mathcal{I}$), a question ($\mathcal{T}$), and a prompt specifying the image modality ($\mathcal{L}$).

As shown in the left panel of Figure~\ref{fig:procedure}, the input image is processed through parallel pathways: a Vision Encoder generates image-level embedding ($E_I$), while the \textbf{RGMo} module jointly processes $\mathcal{I}$ and $\mathcal{T}$ to produce the ROI related embedding ($E_{ROI}$). Simultaneously, the modality prompt, along with $E_{I}$, is passed through a \textbf{TPE} module to yield an enhanced text embedding ($\mathcal{M}$).

The embeddings $E_I$ and $E_{ROI}$, both processed by a Linear Align layer, together with $\mathcal{M}$ and $\mathcal{T}$, are subsequently passed to a Small LLM. This LLM serves as the core reasoning engine, integrating multi-modal inputs to produce a precise textual answer, such as identifying the primary organ in the image.

\subsection{Text Prompt Enhancer (TPE)}

To address the inherent limitations of smaller VLMs, whose reduced parameter size can hinder deep reasoning and lead to attention on irrelevant regions, we introduce the TPE module. This module enriches the input prompt with salient, high-level semantic information from the image, compensating for the model's limited capacity.

The TPE module is built on a pre-trained CLIP model, sharing the vision encoder with our medical VLM and incorporating a single BERT model as the text encoder, thus keeping the parameter overhead low. Given a medical image $\mathcal{I}$, TPE leverages CLIP’s zero-shot classification ability to perform an initial analysis, such as identifying the image modality (e.g., CT, MRI, or X-ray). This classification is then used as a prompt for the VLM’s LLM, providing foundational knowledge that guides reasoning and enables more accurate, context-aware responses, effectively compensating for the limitations of a smaller model.

Specifically, we construct a text  list $\mathcal{L}$, representing possible image modalities, as input to the BERT model in the TPE module to generate the embedding $E_{\mathcal{L}}$. The most relevant text embedding (i.e., the most probable modality) $\mathcal{M}$ is then formally defined as:
\begin{equation}\label{eq:0}
\mathcal{M} = \text{argmax}_i\frac{E_I\cdot E_{\mathcal{L}_i}}{\left \| E_I \right \| \left \| E_{\mathcal{L}_i} \right \|}
\end{equation}
where $E_{\mathcal{L}_i}$ denotes the embedding of the $i$-th element in $\mathcal{L}$.

\subsection{ROI Generation Module (RGMo)}
In general domains, visual prompts significantly improve a model's ability to analyze images by providing crucial spatial reasoning. This is especially beneficial in the medical field, where problem-solving hinges on specific details. However, adapting general-purpose tools like YOLO or Grounding DINO to medical imaging is challenging due to domain shifts, which necessitate costly and time-consuming fine-tuning. This process requires datasets with ground-truth masks of illnesses or organs, which are scarce and expensive to annotate.
To this end, and inspired by the visual explanation tools such as GradCAM, we propose a training-free CLIP-based RGMo module that provides useful bounding boxes without requiring additional training. 

Specifically, we first utilize  the CLIP-based Multi-modal Information Bottleneck ($MI$) module  to generate a saliency heatmap $\mathcal{SA}_{ori}$ from the input $\mathcal{I}$ and $\mathcal{T}$:
\begin{equation}\label{eq:1}
\mathcal{SA}_{ori} = MI(\mathcal{I},\mathcal{T};\theta)
\end{equation}
where $\theta$ represents the parameters of the $MI$ module \cite{m2ib}.
Notably, as shown on the right panel of Figure~\ref{fig:procedure}, while the heatmap's most intense area, or called the fixed point, can remain constant regardless of the input text, we observed that the overall heatmap pattern changes significantly in response to the input text. To leverage this text-dependent behavior for better ROI performance and sharper attention, we concurrently employ a KeyBERT based LLM (denoted as KB) to extract keywords and derive the background text $\mathcal{T}_{back}$ by subtracting them from $\mathcal{T}$:
\begin{equation}\label{eq:2}
\mathcal{T}_{back} = \mathcal{T} - \text{KB}(\mathcal{T})
\end{equation}
which is then processed as in Eq.~\ref{eq:1} to obtain $\mathcal{SA}_{back}$. 

Given that the actual region of interest can align with a fixed point, we adopt the following $S^{3}$ mechanism rather than simply subtracting these two saliency maps:
\begin{equation}\label{eq:3}
\mathcal{SA}_{i,j} = \left\{\begin{matrix} 
  \mathcal{SA}_{ori_{i,j}}-\mathcal{SA}_{back_{i,j}}\text{, if }\mathcal{SA}_{back_{i,j}}\le \delta  \\  
\varepsilon(\mathcal{SA}_{ori_{i,j}}-\mathcal{SA}_{back_{i,j}}) \text{, if }\mathcal{SA}_{back_{i,j}}> \delta
\end{matrix}\right.
\end{equation}

\begin{table*}[]
\caption{Performance comparison with prior state-of-the-art methods on VQA-RAD, SLAKE, and PMC-VQA datasets. Params denotes the total parameter size of the model. Unless marked with $^*$ (indicating results from our reproduction), all other data are taken from the respective original papers. \textbf{Bold} highlights the best performance under each setting.}
\centering
\label{tab:1}
\begin{tabular}{c|c|cc|cc|c}
\hline
                                    &                          & \multicolumn{2}{c|}{\textbf{VQA-RAD}} & \multicolumn{2}{c|}{\textbf{SLAKE}}                                         & \textbf{PMC-VQA}                              \\
\multirow{-2}{*}{\textbf{Models}}            & \multirow{-2}{*}{\textbf{Params}} & Open         & Closed        & Open                        & Closed                               & Accuracy                             \\ \hline
{FAVP (Vicuna)\cite{FAVP}} & 8.5B                     & 71.9         & \textbf{88.2}          & 87.2                        & 88.1                                 & 30.6$^*$                                 \\
{PubMedCLIP\cite{pubmedclip}}   & 1.6B                     & 60.1         & 80.0            & 78.4                        & 82.5                                 & 34.5                                 \\
{LLaVA-Med\cite{llavamed}}    & 7B                   & 64.8         & 83.1          & 87.1                        & 86.8                                 & 42.8$^*$                                 \\
{MedVInT-TD\cite{pmcvqa}}   & 8.6B                     & \textbf{73.7}         & 86.8          & 84.5                        & 86.3                                 & 40.3                                 \\
{MedVInT-TE\cite{pmcvqa}}   & 8.6B                     & 69.3         & 84.2          & \textbf{88.2}               & 87.7                                 & 39.2                                 \\
{PMC-CLIP\cite{pmc-clip}}     & 1.6B                     & 52.0           & 75.4          & 72.7                        & 80.0                                   & 24.0                                 \\
\rowcolor[HTML]{C0C0C0} 
{RoiMAM}         & 1.7B                     &      56.3        &      76.8         & {86.5} & {\textbf{88.2}} & {\textbf{44.9}} \\ \hline
\end{tabular}
\end{table*}

\begin{table}[!htb]
\centering\label{table222}
\caption{Ablation study on the effectiveness of TPE and RGMo. "w/o" means the model without corresponding module. Performance Degration indicates the degrade on performance comparing to original model.}
\begin{tabular}{c|c|c}
\hline
\multicolumn{1}{l|}{\textbf{METHODS}} & \multicolumn{1}{l|}{\textbf{PMC-VQA}} & \multicolumn{1}{l}{\textbf{Performance Degration}} \\ \hline
\multicolumn{1}{c|}{RoiMAM}    & 44.9                                  & 0.0                                                  \\
w/o RGMo                       &    37.0                                   &   7.9                                                 \\
w/o TPE                        &       40.0                                &      4.9                                              \\ \hline
\end{tabular}
\end{table}

\noindent Here, $\mathcal{SA}{i,j}$ denotes a point on the saliency map, $\delta$ is a hyperparameter for selecting high values in $\mathcal{SA}_{back}$, and $\varepsilon$ amplifies suppressed values at fixed points. This strategy focuses on selectively enhancing or suppressing visual attention based on semantic relevance, thereby enabling targeted attention control without additional training.

Once the saliency map $\mathcal{SA}$ is obtained, we extract the ROI image $\mathcal{I}_{ROI}$ via OpenCV’s connected-component analysis. The detected regions are then highlighted with red bounding boxes before the processed image is fed into the vision encoder to compute its embedding $E_{ROI}$.




\section{EXPERIMENT RESULTS AND ANALYSIS}
\label{sec:exp}
\subsection{Experiment Setup}

\textbf{Base Model.} We utilized UniMedCLIP as the base CLIP model for both TPE and RGMo, adopted its vision encoder for RoiMAM, and selected Qwen2-1.5B as the base LLM due to its open-source availability and suitable parameter size \cite{unimed}\cite{qwen2}. Regarding the KeyBERT module, we employed a pretrained all-MiniLM-L6-v2 on PubMedAKE dataset, a dataset for medical keyword extraction \cite{pubmedake}. All the experiments were implemented using PyTorch on a remote server with NVIDIA A100 GPU (40GB memory) and Intel Xeon Gold 5320 (26 cores).

\noindent \textbf{Datasets and Evaluation Metrics.} We fine-tune and evaluate RoiMAM on three MedVQA datasets: VQA-RAD, SLAKE, and PMC-VQA \cite{vqarad}\cite{slake}\cite{pmcvqa}. For PMC-VQA, we report accuracy on multiple-choice questions. Since SLAKE and VQA-RAD each contain open- and closed-set questions, we measure recall for the open set and accuracy—whether the ground-truth label appears in the model output—for the closed set. 

\noindent \textbf{Training Stages.} 
We adopted a two-stage training strategy. Throughout all the training stages, all parameters of our RoiMAM model remain frozen except for the weights of the Linear Align layer and the Small LLM’s low-rank adaptation layers.



In the first stage, regarding differenct downstream tasks, RoiMAM was pretrained on two different datasets, in order to allow RoiMAM have the basic medical knowledge. For downstream tasks on PMC-VQA and SLAKE dataset, we pretrained RoiMAM using the PMC-VQA dataset to produce the answers of the given questions based on the input images. For downstream tasks on VQA-RAD dataset, we pretrained RoiMAM on the radiology subset of the ROCO dataset to produce the original captions for input images, guided by a caption-generation prompt \cite{roco}. During this stage, TPE and RGMo are not involved in the computation and stay inactive.

In the second stage, RoiMAM is fined-tuned and evaluated on respective downstream tasks such as the SLAKE and VQA-RAD datasets, with both TPE and RGMo modules involved in the computation.

\subsection{Quantitative Evaluations}
\label{sec:quantitative evaluations}

Table 1 compares RoiMAM with prior state-of-the-art methods on three MedVQA benchmarks—VQA-RAD, SLAKE, and PMC-VQA—while also reporting each model’s parameter count. RoiMAM uses only 1.7B parameters, far fewer than large competitors such as FAVP (8.5B), MedVInT-TD (8.6B), or LLaVA-Med (7B). Despite its compact size, RoiMAM delivers highly competitive or superior results.

As indicated in Table 1, for the VQA-RAD task, three smaller-scale models show relatively poor performance compared to the four larger-scale ones, which may be due to their limited generalization ability. Our model achieves moderate performance among the evaluated methods, with an open-set recall of 56.3\% and a closed-set accuracy of 76.8\%. Its results are comparable to those of two other smaller-scale models. On the SLAKE dataset, RoiMAM achieves the highest accuracy of 88.2\% in closed-set evaluation, slightly outperforming FAVP (88.1\%) and MedViN-T-TE (87.7\%), despite the latter having approximately five times as many parameters. Furthermore, it reaches 86.5\% recall in the open-set evaluation, substantially surpassing PubMedCLIP (78.4\%) and PMC-CLIP (72.7\%), which have a similar number of parameters. For the PMC-VQA case, RoiMAM achieves 44.9\% accuracy—the highest among all methods—surpassing the next best, LLaVA-Med (42.8\%), by more than two percentage points despite using only a quarter of the parameters. Even more strikingly, compared to PMC-CLIP, our model has only 0.1B more parameters but achieves roughly twice the accuracy.

Overall, Table 1 shows that RoiMAM attains state-of-the-art or near-state-of-the-art results on every benchmark while employing only a fraction of the parameters demanded by other large-scale medical VLMs. This combination of strong performance and compact architecture underscores the model’s efficiency and robustness, rendering it particularly appropriate for clinical or research environments where computational resources are limited.

\subsection{Ablation Study}
During fine-tuning on downstream datasets, the TPE and RGMo modules enable integration of ROI information into the medical images and enrichment with knowledge-rich text prompts, guiding the model to focus on relevant regions rather than extraneous objects. To assess the contributions of TPE and RGMo, we conduct an ablation study on the clean test set of the PMC-VQA dataset in a multiple-choice setting, evaluating performance with or without each module. 
The detailed results are presented in Table 2. 

As reported in Table 2, removing RGMo results in the largest performance drop, with accuracy decreasing from 44.9\% to 37.0\% (a 7.9-point degradation), highlighting RGMo’s critical role in guiding region-focused reasoning. Accordingly, excluding TPE reduces accuracy by 4.9\%, confirming that enriched textual prompts further improve model performance. These results demonstrate that both modules are essential for achieving the full capability of RoiMAM, with RGMo providing the more significant impact. Our experimental findings also confirm that TPE and RGMo play a critical role in boosting MedVQA performance.

\section{CONCLUSION}
\label{sec:con}
In this paper, we introduce RoiMAM, a medical VLM that incorporates two key modules—the ROI Generation Module and the Text Prompt Enhancer—to guide the model toward relevant image regions for MedVQA tasks. The Text Prompt Enhancer leverages a CLIP model, sharing its vision encoder with RoiMAM, and a lightweight BERT layer to generate informative text prompts for the LLM. Accordingly, the ROI Generation Module produces region-of-interest information using a Multi-modal Information Bottleneck that shares the same CLIP backbone. Extensive experiments show that RoiMAM achieves comparable or even state-of-the-art results on MedVQA datasets while using only 20\% of the parameters of leading methods, demonstrating both the effectiveness and efficiency of our model for MedVQA tasks.


\vfill\pagebreak
\bibliographystyle{IEEEbib}
\bibliography{refs}

\end{document}